\documentclass[journal]{IEEEtran}
\usepackage{cite}
\usepackage{amsmath,amssymb,amsfonts}
\usepackage{graphicx}
\usepackage{textcomp}
\usepackage{xcolor}
\def\BibTeX{{\rm B\kern-.05em{\sc i\kern-.025em b}\kern-.08em
    T\kern-.1667em\lower.7ex\hbox{E}\kern-.125emX}}
\usepackage[utf8]{inputenc} 
\usepackage{epstopdf}  
\usepackage{makecell}  
\usepackage[
  separate-uncertainty = true,
  multi-part-units = repeat
]{siunitx}
\usepackage{bm}        

\DeclareMathOperator*{\argmin}{arg\,min}
\newcommand{\e}[1]{\ensuremath{\cdot 10^{#1}}}  
\newcommand*{\tran}{^{\mkern-1.5mu\mathsf{T}}}  

\newcommand{\figref}[1]{\figurename~\ref{#1}}
\newcommand{\tabref}[1]{Tab.~\ref{#1}}

\newcommand{\secref}[1]{Sec.~\ref{#1}}
\begin{document}

\title{Data-Driven Permanent Magnet Temperature Estimation in Synchronous Motors with\\ Supervised Machine Learning
\thanks{This work is supported by the Deutsche Forschungsgemeinschaft (DFG) under grant BO 2535/15-1.}
}

\author{
	\vskip 1em
	{
	Wilhelm Kirchgässner, \emph{Member}, \emph{IEEE},
	Oliver Wallscheid, \emph{Member}, \emph{IEEE}
	\\ and Joachim Böcker, \emph{Senior Member}, \emph{IEEE}
	}

	\thanks{
		
		{
		
		W. Kirchgässner, O. Wallscheid and J. Böcker are with the Department Power Electronics and Electrical Drives, Paderborn University, D-33095 Paderborn, Germany (e-mail: \{kirchgaessner, wallscheid, boecker\}@lea.upb.de).
		}
	}
}

\maketitle

\begin{abstract}
Monitoring the magnet temperature in permanent magnet synchronous motors (PMSMs) for automotive applications is a challenging task for several decades now, as signal injection or sensor-based methods still prove unfeasible in a commercial context.
Overheating results in severe motor deterioration and is thus of high concern for the machine's control strategy and its design.
Lack of precise temperature estimations leads to lesser device utilization and higher material cost.
In this work, several machine learning (ML) models are empirically evaluated on their estimation accuracy for the task of predicting latent high-dynamic magnet temperature profiles. 
The range of selected algorithms covers as diverse approaches as possible with ordinary and weighted least squares, support vector regression, $k$-nearest neighbors, randomized trees and neural networks.
Having test bench data available, it is shown that ML approaches relying merely on collected data meet the estimation performance of classical thermal models built on thermodynamic theory, yet not all kinds of models render efficient use of large datasets or sufficient modeling capacities.
Especially linear regression and simple feed-forward neural networks with optimized hyperparameters mark strong predictive quality at low to moderate model sizes.

\end{abstract}


\section{Introduction}
\label{sec:intro}
\IEEEPARstart{T}HE permanent magnet synchronous motor (PMSM) is the preferred choice in many industry applications due to its high power and torque densities along its high efficiency \cite{ZhuHowe2007}.
In order to exploit the motor's maximum utilization, high thermal stress on the motor's potentially failing components must be taken into account when designing the motor or determining its control strategy.
Especially in the automotive sector, competitive pressure and high manufacturing costs drive engineers to find more and more ways to reduce the safety margin in embedded materials.
Being able to exploit the motor's full capabilities makes precise temperature information at runtime necessary since overheating will result in severe motor deterioration.
Among the typical important components that are sensitive to excessive heat, e.g. stator end windings and bearings, the permanent magnets in the rotor constitute especially failure prone parts of the motor.
Cooling of the rotor is an intricate endeavor compared to stator cooling, which adds to the risk of permanent magnets irreversibly demagnetize due to overheating \cite{Huger2015}.
While sensor-based measurements would yield fast and accurate knowledge about the machine's thermal state, assessing the rotor temperature in this manner is usually not within economic and technically feasible boundaries yet.
In particular, direct rotor monitoring techniques such as infrared thermography \cite{Ganchev2010, Stipetic2012} or classic thermocouples with shaft-mounted slip-rings \cite{Mejuto2010} fall short of entering industrial series production.

Consequently, research focus centers on estimating rotor temperatures, and those of permanent magnets in particular, on a model basis.
Although computational fluid dynamics (CFD) and heat equation finite element analysis (FEA) enjoy good reputation for their rigorous modeling capacities \cite{BoVaSta09}, their high computational demand excludes them from real-time monitoring upfront.
An alternative real-time capable thermal model, called lumped-parameter thermal network (LPTN), approximates the heat transfer process with equivalent circuit diagrams. 
Being partly based on basic formulations of heat transfer theory, they are computationally lightweight if reduced to a low-order structure and provide good estimation performance \cite{WaBo2016}.
However, LPTNs must forfeit physical interpretability of its structure and parameter values by significantly curtailing degrees of freedom in favor of the real-time requirement.
Moreover, expert domain knowledge is mandatory for the correct choice of not only their parameter values, but also for their structural design \cite{WaBo2016}. 

In the last decades, research efforts have also been made that deviate from thermodynamic theory: 
Typical lightweight approaches from this domain encompass the setup of electric machine models that provide information about temperature-sensitive electrical model parameters indirectly.
There are methods that work with current injection \cite{WiSte2010} or voltage injection \cite{Ganchev2011} to obtain the stator winding resistance or the magnetization level of the magnets, respectively, as thermal indicators at the cost of additional losses.
Moreover, fundamental wave flux observers \cite{SpeWaBo2014} can be contemplated to assess the reversible demagnetization of the embedded magnets.
However, these methods suffer from high electric model parameter sensitivity, such that inaccurate modeling (potentially in the range of manufacturing tolerances) leads to excessive estimation errors \cite{WaHuPe14}.

In an effort to combine the advantages from both domains, \cite{Gaona2020} fused an LPTN and a flux-based temperature observer with a Kalman filter.
They report increased robustness and estimation accuracy for the full motor speed range, and an additional system failure detection feature.

In contrast to these physically-motivated estimation approaches, machine learning (ML) models that detach from any classic fundamental heat theory approximation will be examined empirically on the task of estimating the magnet temperature in a PMSM in this work.
No a priori knowledge will be incorporated, and computational demand at runtime (during inference) scales less drastically with model complexity.
Model parameters are fitted on observational data only, making domain knowledge less relevant.
Leveraging this generalizability, one can easily transfer insights from this work into neighboring fields of interest, e.g. heating in power electronics, batteries' state-of-health, etc.

\begin{figure}
    \centering
    \includegraphics[width=0.49\textwidth]{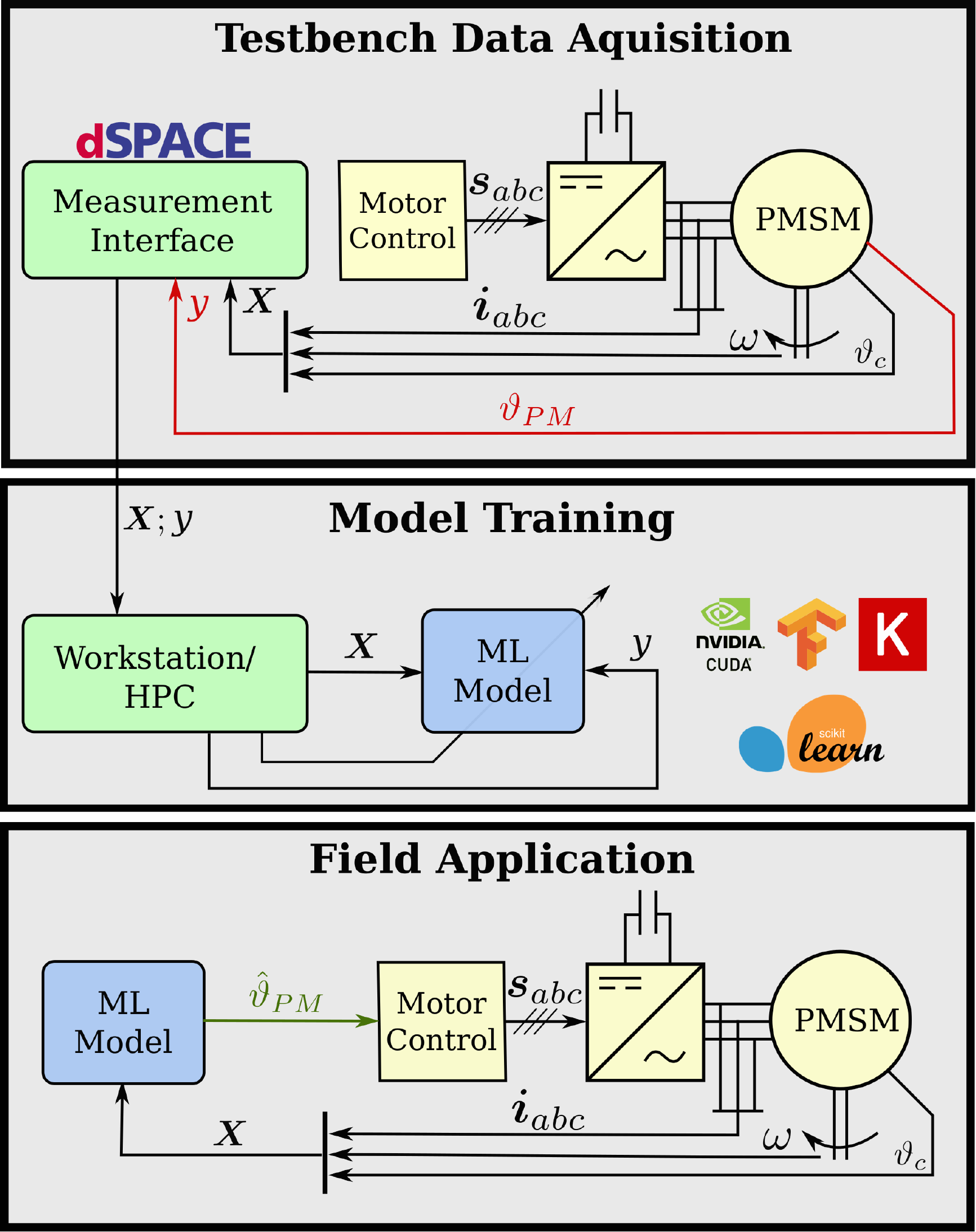}
    \caption{Simplified scheme of the whole process from data acquisition at the test bench over model training up to the integrated temperature monitoring in an automotive application (cf. \cite{KiWaBo2019}).}
    \label{fig:scheme}
\end{figure}

A scheme depicting the idea of fitting a ML model on collected testbench data and having it eventually inform an arbitrary controller is shown in \figref{fig:scheme}.
The more accurate the control is informed of the thermal state, the better it can watch for critical operation and apply power derating \cite{WaBo2017}.

Although the magnet temperature is the only contemplated target value in this work, all considered ML models are also trivially applicable to other continuous or discrete valued quantities of interest, such as torque or stator temperatures \cite{WaKiBo17, KiWaBo2019, KiWaBo2019_2}.
Furthermore, incorporating an increasing amount of spatially targeted temperature regions poses a virtually minor design overhead.

\subsection*{Related Work}
Certain ML approaches for the task of temperature profile estimation in a PMSM were studied before:
Recurrent neural networks with memory units, in particular, long short-term memory (LSTM) or gated recurrent units (GRU) were evaluated on low-dynamic temperature profiles with a hyperparameter optimization via particle swarm optimization (PSO) in \cite{WaKiBo17}.
In \cite{KiWaBo2019}, temporal convolutional neural networks (TCN) were applied on also high-dynamic data and a comparison with recurrent architectures were compiled after tuning hyperparameters with Bayesian optimization.
Far simpler ML models like linear regression were also shown to be effective as long as data has been preprocessed with low-pass filters in \cite{KiWaBo2019_2}.

This paper extends the related work in so far that the broader field of supervised learning in regression tasks is illuminated. 
All previous publications fall into the regime of either fast and simple least-squares regression, or computation-heavy and sophisticated deep learning, albeit there is a rich set of tools in between.
This gap is characterized by models of intermediate complexity and expressive power, and is systematically evaluated in this work with real-time capability and achievable estimation accuracy in mind.

All experiments are conducted on the dataset\footnote{Publicly available at www.kaggle.com/wkirgsn/electric-motor-temperature} from \cite{KiWaBo2019, KiWaBo2019_2}.

\section{Regression Algorithms}
\label{sec:algos}
The range of ML models can be categorized by e.g. their modeling capacities, prior assumptions over the data, amount of model parameters and their update rules, or simply their runtime in terms of the big-$\mathcal{O}$ notation. 
Representative classes for the regression task are linear models like ordinary least squares and its regularized derivatives; feed-forward neural networks as non-linear function approximators; decision trees that learn splits in the feature co-domain; support vector machines; k-nearest-neighbors; as well as the diverse ensemble of those (cf. \cite{Hastie2009}).
All contemplated models are briefly illuminated in this section.

\subsection{Ordinary Least Squares}
The model family of linear approximators assume a linear relationship between a multi-dimensional input of vectors $\bm{X} = (\bm{x}_1, \bm{x}_2, ..., \bm{x}_p) \in \mathbb{R}^{n\times p}$ and the real-valued (possibly multi-dimensional) output vector $\bm y$ with $p$ denoting the amount of input features, and $n$ being the amount of obervations.
See \cite{Hastie2009} for a comprehensive overview.

Rearranging the minimization of the residual sum of squares gives a closed solution form for inferring the model coefficients $\bm{ \hat \beta}$ from the data:
\begin{equation}
	\label{eq:lin_model}
    \bm{\hat y} = \bm{X \hat \beta} = \bm X(\bm X\tran\bm X)^{-1}\bm X\tran\bm y,
\end{equation}
which is known as ordinary least squares (OLS).
Popular regularized versions of OLS are called ridge regression \cite{HoeKen1970} for an additional $\ell_ 2$ penalty term in the cost function or LASSO \cite{Tibshirani1996} for the $\ell_1$ addition.
Here, these are not considered as they have been shown to be inefficient for this dataset \cite{KiWaBo2019_2}.

\subsection{Weighted Least Squares}
A variation of OLS where all observations are weighted by a weight matrix $\bm W \in \mathbb R^{n\times n}$:
\begin{equation}
    \bm{\hat \beta} = (\bm X\tran\bm W\bm X)^{-1}\bm X\tran\bm W\bm y,
\end{equation}
Weighted least squares (WLS) is often used to account for heteroscedasticity in the data and increase robustness of the estimator. 
Especially in this work, there is another industry-driven incentive to weight observed data: Since the ultimate goal is to avoid overheating, it is of reasonably higher interest to estimate high temperatures more accurately than lower thermal states.
When deviating from the analytical solution of the least squares method to gradient-descent-based optimization, one can also penalize under-estimates more than over-estimates, which is not covered here.
WLS will be compared to OLS in Sec.~\ref{sec:err_residuals}.

\subsection{Epsilon-Support Vector Regression}
Although observations $\bm x \in \mathbb{R}^p$ are usually projected into higher dimensions through a kernel function $K(\bm x, \bm{x'}) = \langle \phi(\bm x), \phi(\bm x')\rangle$, epsilon-support vector regression ($\epsilon$-SVR) still constructs a regularized linear model with coefficients $\bm \beta$ on the new feature space \cite{Hastie2009}.
Here, regularization means penalizing model complexity by minimizing the quadratic weights.
More specifically, the linear cost function is $\epsilon$-insensitive i.e. cost is accumulated only if the prediction error exceeds a threshold $\epsilon$.
Those observations with errors beyond $\epsilon$ are called support vectors in the regression context.
In order to allow for support vector deviation from the $\epsilon$-band one can encompass non-negative slack-variables $\bm\xi$ and $\bm\xi'$ in the minimization formulation:
\begin{equation}
		\label{eq:svm_opt}
        \argmin_{\bm\beta} \Big(\frac{\bm\beta\tran\bm\beta}{2} + C\sum_{i=1}^{N} (\xi_i + \xi'_i)\Big),
\end{equation}
\begin{align*}
       \text{s.t.}\quad\quad y_i - \hat{y_i} &\leq \epsilon + \xi'_i , \\
       \hat{y_i} - y_i &\leq \epsilon + \xi_i,\\
       \xi_i,\xi'_i &\geq 0\quad \forall\quad i \in \{1\, ..\, \mathrm N\},
\end{align*} 
where $C$ balances coefficient regularization and $\epsilon$-band alleviation.
Solving the dual problem gives the following approximate solution:
\begin{equation*}
    \bm{\hat{y}} = \sum_{i=1}^M (\alpha_i - \alpha'_i)K(\bm{x}_i,\bm{x})\quad 
    \text{s.t.}\quad 0 \leq \alpha'_i \leq C, 0 \leq \alpha_i \leq C,
\end{equation*}
with $M$ being the number of support vectors, and $(\alpha_i - \alpha'_i)$ denoting the weight for support vector $i$.


\subsection{\textit{K}-Nearest Neighbors}
In the regression context, $k$-nearest neighbors ($k$-NN) is a method where all training observations are stored, while new samples are estimated by taking the mean of stored points in the vicinity of that new observation \cite{Hastie2009}.
The design parameter $k$ is the number of neighbors to consider when evaluating the mean.
Neighborhood is determined by the euclidean norm, and can be weighted optionally by the distance of each neighbor to the new sample.
The $k$-NNs denote a so-called lazy learn algorithm, since all computation is deferred to the inference phase.
This counteracts real-time capability but might be bought for superior accuracy.

\subsection{Randomized Trees}
Two methods based on ensembling of randomly grown decision trees are included: Random forests (RF)\cite{Breiman2001} and extremely randomized trees (ET)  \cite{Geurts2006}.
Single decision trees tend to overfit on the data and, therefore, suffer from high variance.
This is mitigated by building an ensemble of decision trees that are fit on random subsets of the given observations with replacement (bootstrapping) and random subsets of features.
Both measures lead to diverse predictions that are partially decorrelated from each other, such that averaging over them reduces variance significantly at the cost of additional bias.
ETs differ from RFs in so far that, during training, ETs draw random thresholds in each feature out of the considered set in order to determine the next split, whereas RFs search for the most discriminative thresholds.
This additional random component in ETs further amplifies the effect of variance reduction and bias increase.

\subsection{Neural Network Architectures}
Neural networks are known to be universal generalizers \cite{HoSti1989}, with gradual degrees of complexity that can be adapted to the capacities of the application platform.
The vanilla form of a neural network is the multi-layer perceptron (MLP), where regressors are non-linearly transformed over several layers and eventually conclude to a prediction through (\ref{eq:lin_model}).
The transformed vector $\bm h^{(l)}$ after layer $l$ is computed from the preceding transformed vectors of layer $l-1$ by
\begin{equation}
    \bm h^{(l)} = g^{(l)}(\bm W^{(l)}\bm h^{(l-1)}+\bm b^{(l)});\quad l \in \{1\, .. \, \mathrm L \} 
\end{equation}
with $g^{(l)}$ being an activation function at layer $l$ and $\bm W^{(l)} \in \mathbb R^{r\times s}$ denoting the trainable weights between the $r$ and $s$ neurons of layer $l$ and $l-1$, respectively.
Popular choices for the activation function are the rectified linear unit (ReLU) \cite{Nair2010}
or the exponentially linear unit \cite{CleUntHoch2015}
as they have shown to converge faster with similar accuracy compared to the original sigmoid or tangens hyperbolicus.
Despite the high non-linear structure in MLPs, they are end-to-end differentiable through the backpropagation rule \cite{HoSti1989}, making them optimizable by stochastic gradient descent (SGD) and derivations from that. 

A standard regularization scheme, next to weight decay \cite{MoHa1995} and dropout \cite{SriHiKri2014}, is batch normalization \cite{batchnorm2015}, which scales each new batch of training data after every layer according to the $\ell_2$ norm but has no effect during inference \cite{hoffer2018norm}.
Even though this and alternative normalization schemes, such as weight and layer norm \cite{weightnorm2016, layernorm2016}, gained popularity in recent years \cite{hoffer2018norm}, there was also a new type of activation and dropout functions proposed that circumvent additional layers: Self-normalizing NNs (SNNs) \cite{snn2017}.
Here, the idea is to normalize neuron activations implicitly through scaled ELUs (SELU) and a new variant of dropout.

In contrast to the MLP architecture, recurrent topologies with long short-term memory (LSTM) \cite{GeSch1999} or temporal convolutional neural networks (TCN) \cite{BaiKoKo2018} can utilize time dependency between neighboring observations in a temperature profile without explicit feature engineering.
However, these variants come with many more model parameters, and were extensively optimized in \cite{KiWaBo2019}, such that their performance will be merely reported for comparison with the benchmarks in the end of this paper. 

\section{Black-Box Thermal Modeling}
\begin{figure*}
    \centering
    \includegraphics[width=\textwidth]{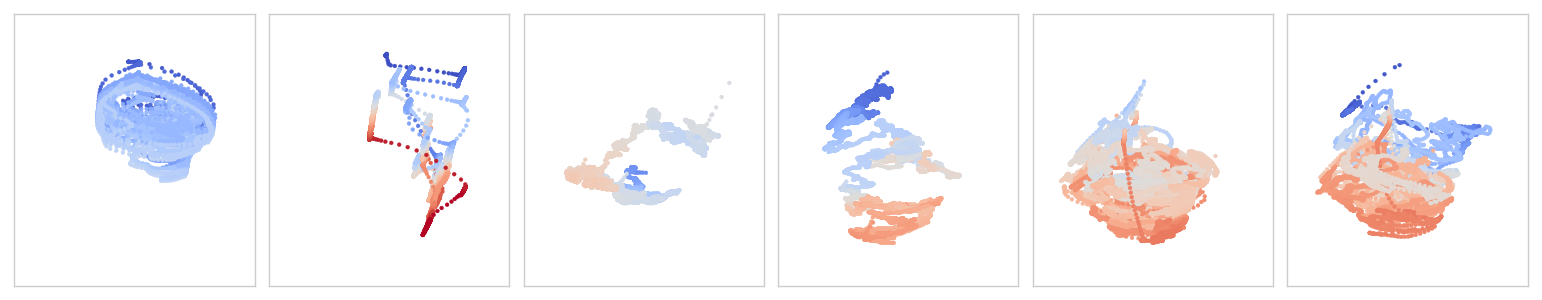}
    \caption{Two-dimensional PCA of the data projected on the first two principal components, that is, the highest sample variance. 
    Several recording sessions are depicted with increasingly random excitation from left to right.
    Those measurements with high PM temperature are colored red and those with low temperature blue. 
    No clear trend is visibly inferrable especially in the middle region. }
    \label{fig:2dPCA}
\end{figure*}

Similar to \cite{WaBo2016}, a three-phase PMSM of \SI{50}{\kilo\watt} mounted on a test bench yielded the available data with a consistent sampling frequency of $f_s = \SI{2}{\hertz}$. 
The data aggregates 139 hours of recordings in total or around one million multi-dimensional samples.
Obviously, supervised learning requires data measured on enhanced motor test equipment.
Coolant, ambient, and magnet temperatures are recorded with standard thermocouples.
The rotor temperature information, represented by the permanent magnets' surface temperature, is transmitted wirelessly over a telemetry unit. \tabref{tab:params} compiles the considered quantities that represent the input and output of the following ML models.
Denoted input signals are commonly accessible in real-world traction drive systems, hence, tuned ML models can be plugged into commercial vehicle controls without further sensor upgrades.
\figref{fig:2dPCA} depicts a two-dimensional principal component analysis (PCA) representation of the input features colored according to the target's thermal state.
It becomes evident that no trivial relationship between data with high and low temperatures is inferable.

\begin{table}[tbp]
    \centering
    \caption{Measured input and target parameters}
    
    \begin{tabular}{l c}
        \hline\hline
        \thead[l]{Parameter name} & \thead{Symbol} \\
        \hline
        \multicolumn{2}{c}{Inputs}  \\
        Ambient temperature 		  & $\vartheta_a$ \\
        Liquid coolant temperature & $\vartheta_c$ \\
        Actual voltage $d$-axis component & $u_d$ \\
        Actual voltage $q$-axis component & $u_q$ \\
        Actual current $d$-axis component & $i_d$ \\
        Actual current $q$-axis component & $i_q$ \\
        Motor speed & $n_{mech}$\\
        \hline
        \multicolumn{2}{c}{Derived inputs} \\
        Voltage magnitude $\sqrt{u_d^2 + u_q^2}$ & $u_s$ \\
        Current magnitude $\sqrt{i_d^2 + i_q^2}$ & $i_s$ \\
        Electric apparent power $1.5* u_s * i_s$ & $S_{el}$\\
        Motor speed and current interaction & $i_s \cdot \omega$  \\
        Motor speed and power interaction & $S_{el} \cdot \omega$  \\
        
        \hline
        \multicolumn{2}{c}{Target}\\
        Permanent magnet temperature & $\vartheta_{PM}$\\
        \hline\hline
    \end{tabular}
    \label{tab:params}    
\end{table}

\subsection{Data Preprocessing and Feature Engineering}
All representations of the data are standardized on their sample mean and sample unit variance exhibited in the training set. 
The exponentially weighted moving average (EWMA) and standard deviation (EWMS) are taken into account so that for every timestep $t$ the following terms are computed for each input parameter $X$ and adhered to the models' input (see \tabref{tab:params}):
\begin{equation}
    \label{eq_ewma}
    \mu_t = \frac{\sum_{i=0}^{t}w_ix_{t-i}}{\sum_{i=0}^tw_i}
    \quad\mathrm{and}\quad
    \sigma_t = \frac{\sum_{i=0}^tw_i(x_i-\mu_t)^2}{\sum_{i=0}^tw_i},
\end{equation}
where $w_i=(1-\alpha)^i$ with $\alpha=2/(s+1)$ and $s$ being the span that is to be chosen.
Multiple values for the span can be applied leading to different frequency-filtered versions of the raw time-series data.
The weights of the preceding $s$ observations describe around \SI{86.5}{\percent} of all weights' total sum.
Consecutive calculations of the EWMA can be derived from a more computationally efficient form of (\ref{eq_ewma}),
\begin{equation}
	\label{eq_ewma_efficient}
    \mu_t = (1-\alpha)\mu_{t-1} + \alpha x_t, 
\end{equation}
which is highly relevant especially for automotive applications where embedded systems run on cost-optimized hardware.
A computing-efficient form of the EWMS exists likewise.

Incorporating these additional features is inevitable for the ML approaches considered in this work, as they all assume independent and identically distributed data.
This assumption is in conflict with a PMSM's thermal behavior representing a dynamic system.
Nonetheless, adding trend to the actual input space gives rise to overcoming this discrepancy and finding approximating functions with sufficient accuracy.

\subsection{Analogy to LPTN RC Circuits}
Examining LPTNs that are well-defined for the temperature estimation task from \cite{WaBo2016}, reveals that these are characterized by low-pass filters or, equivalently, RC circuits smoothing the raw input data.
From signal theory, it is known that RC circuits are infinite impulse response (IIR) filters of the form
\begin{equation}
	x = y + RC\frac{\text{d}y}{\text{d}t},
\end{equation}
which can be discretized to 
\begin{equation}
	\label{eq_discret_IIR}
	x_t = y_t + RC\frac{y_t-y_{t-1}}{h},
\end{equation}
with $h$ being the step size.
Rearranging (\ref{eq_discret_IIR}) gives
\begin{equation}
	y_t = \frac{RC}{RC+h}y_{t-1} + \frac{h}{RC+h}x_t,
\end{equation}
which resembles (\ref{eq_ewma_efficient}) with $\alpha = (RC + h)^{-1}h$.

Consequently, it is reasonable to directly apply EWMAs on the sensor time-series recordings in order to obtain regressors exhibiting patterns similar to those in LPTNs.
This observation was empirically confirmed in \cite{KiWaBo2019_2}.

\subsection{Cross-Validation}
\label{sec:CV}
Generalization error is reported by evaluating the prediction error on a test set of seven hours unseen during training.
In case of MLPs, further \SI{10}{\percent} of the training set is withheld from training, and acts as validation set.
That portion is used to apply early stopping \cite{deeplearningbook} i.e. mitigating overfitting by stopping training after the cost function on this set is not improving anymore for a certain delta after a given number of iterations (epochs).

Scores are reported upon the mean squared error (MSE), mean absolute error (MAE), and the coefficient of determination (\textit{R}²) between predicted sequence and ground truth.
Moreover, the maximum deviation occuring in the testset ($\ell_\infty$ norm) is an important indicator for the quality of temperature estimations.

\subsection{Cross-Validation for Hyperparameter Tuning}
For a fair comparison of the different models, all design-parameters or hyperparameters are tuned systematically with the same approach: Bayesian optimization \cite{Shahriari2016}.
During this sequential optimization technique, a surrogate model (here, a Gaussian process) is trained to find a mapping from the hyperparameter space to the test set error.
The surrogate model's capability to yield uncertainty estimates in the hyperparameter space can be used to trade exploration off for exploitation when determining the next hyperparameter set to evaluate. 

More specifically, the chosen objective is the average MSE over all folds during stratified group-three-fold cross-validation (CV).
Here, stratification refers to distributing recording sessions with same-level maximum temperatures evenly over the three folds.
Such homogeneous folds reduce test error, following the heuristic of avoiding outlier samples concentrated in just few folds. 
Grouping denotes that no samples from the same measurement session appear in more than one fold in order to mitigate overfitting.
In addition, feature value normalization was conducted for every fold with respect to observations in the training set.
Note that the test set from Sec.~\ref{sec:CV} is not part of this CV strategy, and thus does not leak into the optimization of hyperparameters.

\subsection{Hyperparameters and Intervals}
\begin{table}[tbp]
    \centering
    \caption{Hyperparameter intervals and optimum}
    
    \begin{tabular}{c c c}
        \hline\hline
        \thead[l]{Hyperparameter} & \thead{Interval} & \thead{Optimum} \\
        \hline
        
        \multicolumn{3}{c}{\thead{SVR}}\\
        $C$ & $10^{-3} \dots 10$ (log) & 1.56\\
        $\epsilon$ & $10^{-2} \dots 1$ & 0.11\\
        \hline
        \multicolumn{3}{c}{\thead{$k$-NN}}\\
        neighbors & $1\dots 2048$ & 2048\\
        weighting & uniform or distance & distance\\
        \hline
        \multicolumn{3}{c}{\thead{RF / ET}}  \\
        estimators & $10\dots 600$ & 93 / 600\\
        max. depth & $10\dots 60$ & 60 / 53\\ 
        min. samples for split & $2\dots 20$ & 15 / 20\\
        min. samples per leaf & $1\dots 10$ & 2 / 7\\
        bootstrap & yes or no & yes / yes\\
        \hline
        \multicolumn{3}{c}{\thead{MLP}}\\
        layers & $1\dots 3$ & 1 \\
        units & $4\dots 32$ & 16\\
        activation & SELU or ReLU & ReLu\\
        dropout & $0\dots 0.3$ & 0.13\\
        $\ell_2$-regularization & $10^{-9}\dots 0.1$ (log) & 1.7\e{-8}\\
        learnrate & $10^{-6}\dots 0.1$ (log) & 5.8\e{-3}\\
        optimizer & \makecell{RAdam, Adam, NAdam,\\ Adamax, RMSProp, SGD} & Adam\\
        \hline\hline
        
    \end{tabular}
    \label{tab:hyperparams}    
\end{table}

The certain choice of four span values $\bm s = (s_1, s_2, s_3, s_4)$ was always part of the optimization resulting in an independent span set for each model.
The amount of span values could also be made a hyperparameter, but was deliberately set to four in order to maintain comparability with previous work \cite{KiWaBo2019, KiWaBo2019_2}, where it was found to balance modeling accuracy with computational demand.
Model-specific hyperparameters are compiled in \tabref{tab:hyperparams}.
All hyperparameter interval bounds are chosen manually to a range where modeling performance is likely to converge while constraining runtime. 

RFs and ET come with the same hyperparameters: The number of trees denotes the ensemble size, while the maximum tree depth constraints growth.
A higher minimum amount of samples for a split help make more robust splits, and increasing the minimum amount of samples per tree-leaf can mitigate overfitting.

The MLP model family also offers a wide variety of hyperparameters:
The low upper bounds for the number of units and layers curtail the amount of model parameters and, thus, the modeling flexibility for each MLP, yet it has been shown in \cite{KiWaBo2019} that also smaller neural networks reach satisfactory prediction accuracy.

The particular optimization algorithm was either one of Adam, Adamax, Nesterov Adam (NAdam), rectified Adam (RAdam), RMSprop, or vanilla SGD \cite{KiBa2014, nadam2016, radam2019}.
During training, the learning rate of the optimizer is divided by two after there is no improvement in training set loss for $10$ consecutive epochs anymore. 
Reducing the learning rate on such loss plateaus is a common heuristic in MLP training for improving convergence and local minima exploration \cite{deeplearningbook}.
In addition to this learning rate decay schedule, mini-batch size is doubled after $33$ epochs from $32$ to $64$ and then again to $128$, such that each training lasts $99$ epochs if no early stopping applies.
Increasing the batch size has been shown to benefit training convergence similar to learning rate decay \cite{SmiKiLe2017}.

Hyperparameters of SVR and $k$-NN are described in \secref{sec:algos}.

\section{Experimental Results}
For all hyperparameter optimizations, the scikit-optimize framework \cite{ScikitOpt} is utilized.
The acquisition function is either of upper confidence bound, expected improvement or probability of improvement, which calculate new candidate points independently, and where the most promising proposal serves as next evaluation point in every iteration.
Each model's hyperparameter space was searched for at least 100 iterations with 30 initial random selections.

The Tensorflow-toolbox with its high-level API Keras \cite{keras2015} is used for neural network training, while all other regression algorithms are computed with the scikit-learn toolbox \cite{scikit-learn}.
The libsvm package \cite{ChaLin11} and the radial basis function-kernel are used for the SVR.

The found optima are also organized in \tabref{tab:hyperparams}.

\subsection{Model Performance}
An overview of the individual predictive performance of each model with optimized hyperparameters is compiled in \tabref{tab:model_performance}.
For models with stochastic optimization, the best experiment out of $10$ repetitions with different random number generator seeds is reported in order to alleviate scatter of the training process.
Among the usual performance metrics, the model size is also highlighted.
This quantity represents the amount of parameters that are to be stored for each model in order to make a prediction on a new observation after training.
Note, however, that these numbers do not include memory that must be reserved for saving the moving averages of the sensor data.

\begin{figure*}
    \centering
    \includegraphics[width=\textwidth]{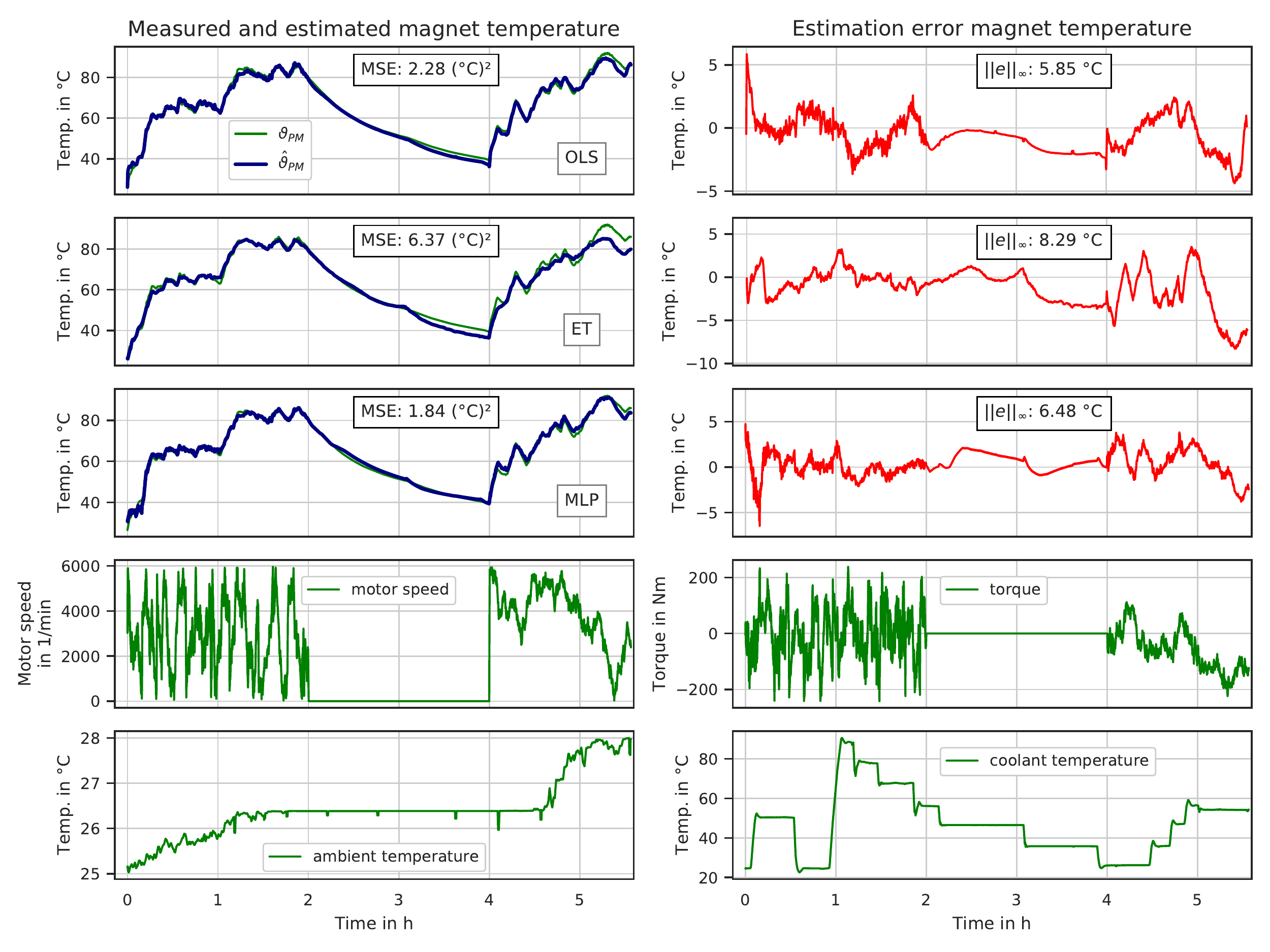}
    \caption{The three best models evaluated on the first test set profile (out of two). Ground truth in green, prediction in blue, and the residual in red. The bottom four plots show selected input variables to highlight the high dynamics of the drive cycle.}
    \label{fig:testset_pred}
\end{figure*}

Black-box CNNs show the overall best performance in terms of the MSE, while grey-box LPTNs have the lowest $\ell_\infty$ norm and merely 46 parameters, which is substantially less than for the other models.
Only OLS is worth mentioning to be in between those two extrema, with a strong MSE, low maximum deviation and also few model parameters of 109.
This should make OLS the preferrable approach among machine learning models in case there are no resources to hand-design an LPTN.
Though deep learning models are the most precise, the model size of over $67$ thousand parameters might be difficult to justify while facing the marginal increase in accuracy especially in automotive systems. 
The test set predictions of OLS, ET and MLP are shown in \figref{fig:testset_pred}.
They can be utilized as intermediate approaches to intelligent temperature estimation before automotive hardware becomes as strong as it is necessary for deep learning.

It becomes evident that $k$-NN, RF and SVR could not find a sufficient function approximation, even though their hyperparameter optimization seeked to maximize modeling capacity by increasing the model size up to the upper bound of the hyperparameter intervals, yet without comparable success.

\begin{table}
\centering
\caption{Benchmark values for different ML methods on the test set}
\begin{tabular}{c | c c c c c c}
        \hline\hline
        Model & \makecell{MSE \\in $\si{\celsius\squared}$} & \makecell{MAE \\in $\si{\celsius}$} & \textit{R}² & \makecell{$\ell_\infty$ norm\\ in $\si{\celsius}$} & model size \\
        \hline
        $k$-NN  & 26.10 & 4.24 & 0.87 & 12.86 & 221k\\
        RF & 16.26 & 3.09 & 0.92 & 10.9 & 1.1M \\
        SVR & 13.42 & 2.75 & 0.93 & 31.99 & 209k \\
        ET & 6.51 & 1.77 & 0.97 & 8.29 & 5.5M \\
        LPTN \cite{GeWaBo2020} & 5.73 & 1.98 & 0.97 & \textbf{6.45} & \textbf{46} \\
        RNN \cite{KiWaBo2019} & 3.26 & 1.29 & 0.98 & 9.1 & 1.9k \\
        MLP & 3.20 & 1.32 & 0.98 & 8.34 & 1.8k \\
        OLS & 3.10 & 1.46 & 0.98 & 7.47 & 109 \\
        CNN \cite{KiWaBo2019} & \textbf{1.52} & \textbf{0.85} & \textbf{0.99} & 7.04 & 67k \\
        \hline\hline
        
\end{tabular}
\label{tab:model_performance}
\end{table}

\subsection{Learn Curves}
\begin{figure}
    \centering
    \includegraphics[width=0.49\textwidth]{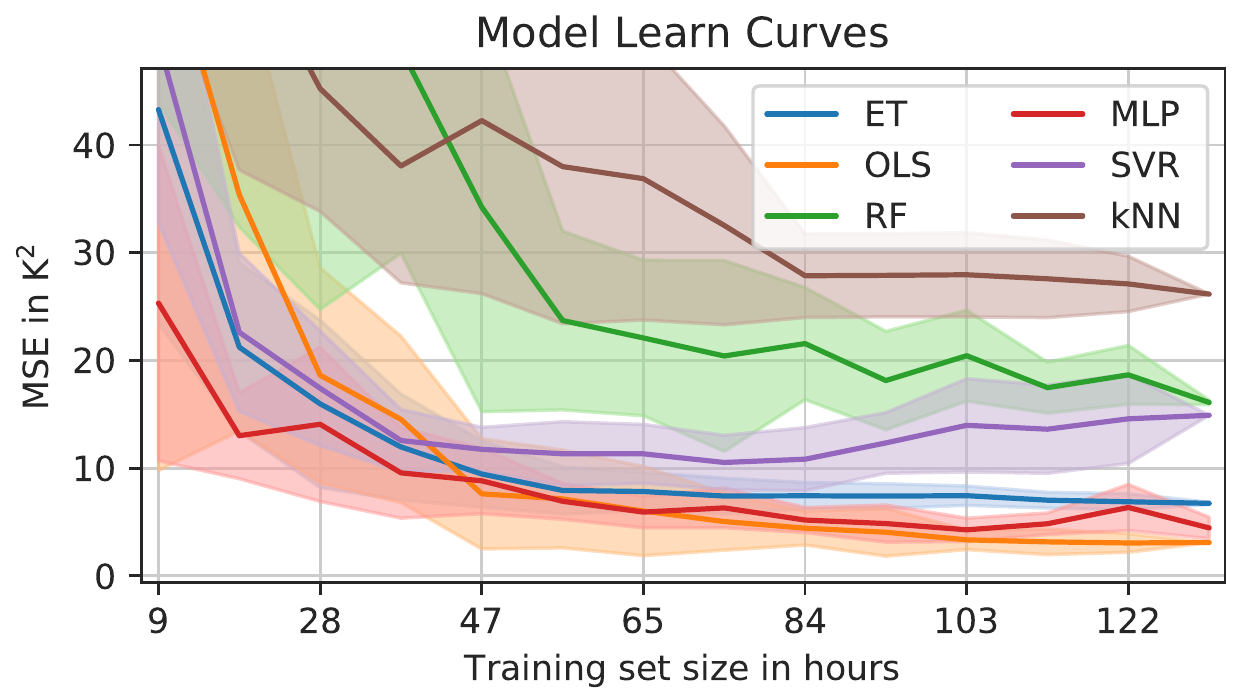}
    \caption{MSE on a fixed test set and an increasing training set size. Lines denote mean MSE among 10 random subsets and transparent area marks standard deviation.}
    \label{fig:learn_curves}
\end{figure}

Besides the total test error of each model, scalability with more training data is often of equal interest.
\figref{fig:learn_curves} illustrates the learn curves of all models.
The test set is constant and the same as in the previous experiments while the training set is increased successively.

It can be seen that all models plateau out at half the training set size except for SVR, whose performance seems to diminish.
An explanation might be a limited modeling capacity for the SVR, which struggles to map all operation points observed in the data.

It can be summarized that the better performing algorithms (OLS, MLP, and ET) achieve high performance already with lesser training set sizes of around $70$ hours and no significant performance gains for more data.
This naturally suggests their use in applications where a limited amount of data is collectable.

\subsection{Error Residuals}
\label{sec:err_residuals}
\begin{figure}
    \centering
    \includegraphics[width=0.49\textwidth]{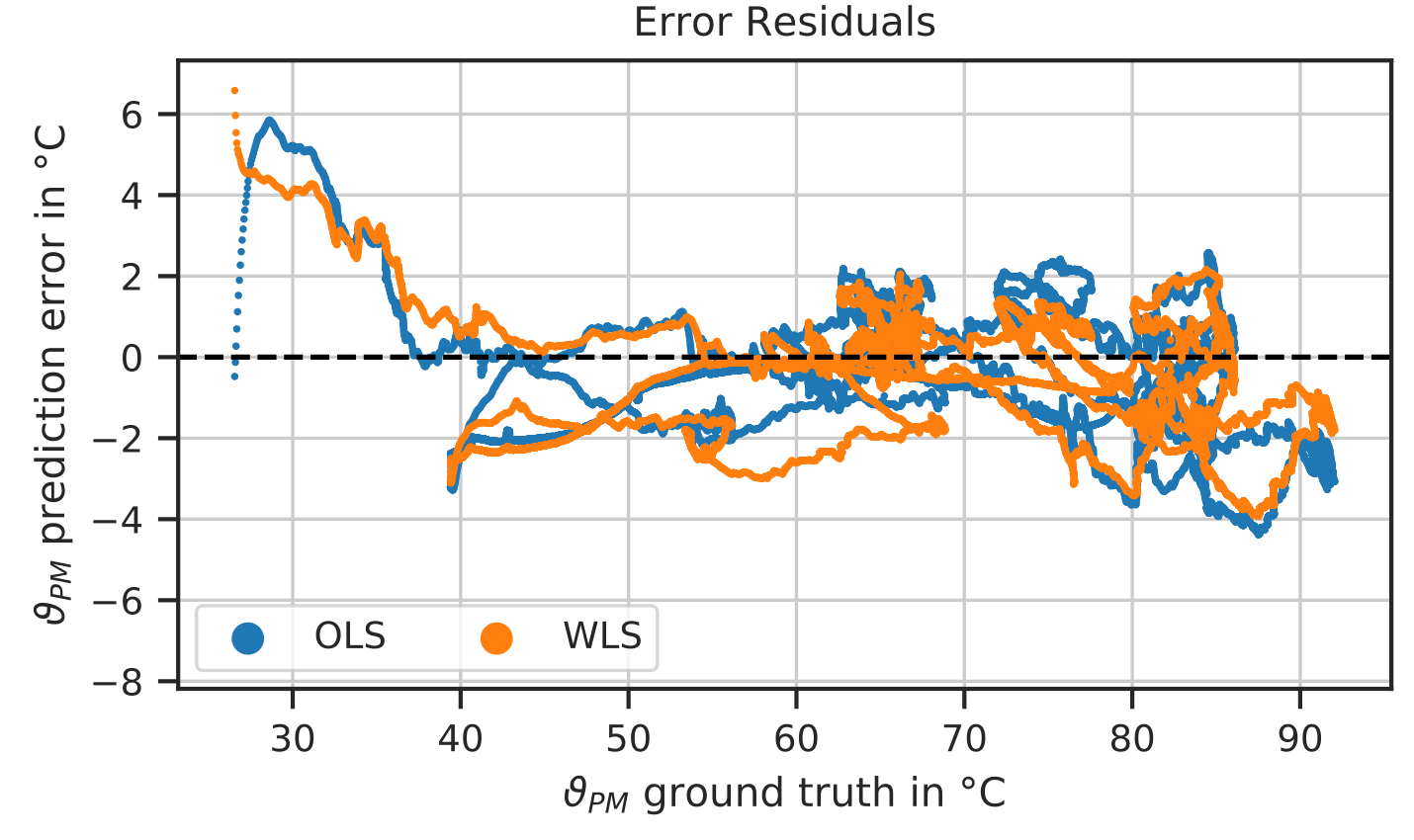}
    \caption{OLS and WLS error residuals on the test set.}
    \label{fig:error_residuals}
\end{figure}

In the following the error residuals along the value range of the ground truth permanent magnet temperature is illuminated.
In terms of an industrial application, robust and accurate estimation of high temperatures is of significantly higher interest than that of low temperatures.
This is due to the purpose of avoiding overheating and the material destruction implied by that.
There are several ways to opt for this more specialized use case:
\begin{itemize}
    \item subsample data such that more high temperatures occur in the data,
    \item adjust the cost function to penalize under-estimates, and increase costs for deviations at high temperatures.
\end{itemize}
While the first point is a general approach to the certain method of data collection and should be studied on its own, the latter point is trivially done for most ML models.
One example could be WLS: their performance and corresponding error residuals are opposed to those of OLS in \figref{fig:error_residuals}.
All obervations are linearly weighted from $0.33$ to $1$ according to their closeness to the total minimum and maximum PM temperature, respectively, occuring in the data set.
The effect is subtle, but it can be seen that for WLS outlying predictions are often settled in lower regions of the temperature value range, while higher temperatures come with less variance.
In order to evaluate the advantage of WLS over OLS on a significant scale, additional load profiles need to be recorded that expose more temperature variance.
Specifically, deviations at low temperatures are inversely, linearly weighted from a maximum weight at the maximum allowed temperature, and under-estimates are additionally weighted by a factor of $\sqrt{10}$. 

\section{Conclusion and Outlook}
It has been shown that if rich datasets are recorded at a test bench or in production, which is likely in the automotive industry, then engineers can rely on them to model temperature estimators.
The utilization of special domain expertise and motor sheet specifications is circumvented, while monitoring important component temperatures inside a PMSM is still real-time processible for certain classical supervised learning algorithms.
Through autonomously conducted hyperparameter searches, it was possible to demonstrate that classical supervised learning algorithms achieve state-of-the-art estimation accuracy also during high dynamic drive cycles.
Ordinary least squares stands out with one of the best accuracies and by far the lowest amount of model parameters among ML methods, making it the first choice after having found optimal moving average factors during feature engineering for this certain application.
In the long run, however, deep neural networks are expected to be prevalent in the temperature estimation domain, due to their excellent scalability, the rising availability of measurement data, and increasing computing capabilities finding their way into series production for the sake of (hybrid) electric vehicles.

It is still an open question how well supervised learning algorithms may generalize across different motors from the same manufacturer or even among different manufacturers.
This can be answered only with a dataset exhibiting this diversity, and is yet to be recorded.
Moreover, incorporating domain knowledge at a lesser scale is an auspicious option.
Eventually, assessing the uncertainty of predictions in order to enable probabilistic estimations could leverage reliability on data-driven temperature estimators.

\bibliography{library.bib}

\end{document}